\documentclass[letterpaper]{article} 
\usepackage{aaai2026}  
\usepackage{times}  
\usepackage{helvet}  
\usepackage{courier}  
\usepackage[hyphens]{url}  
\usepackage{graphicx} 
\usepackage{natbib}  
\usepackage{caption} 
\usepackage{algorithm}
\usepackage{algorithmic}
\usepackage{tikz}
\usepackage{booktabs}
\usepackage{amsmath,amssymb}
\usepackage{mathtools} 
\usepackage[hidelinks]{hyperref}
\usepackage[capitalise,noabbrev]{cleveref}
\usepackage{newfloat}
\usepackage{listings}
\DeclareCaptionStyle{ruled}{labelfont=normalfont,labelsep=colon,strut=off} 
\floatstyle{ruled}
\newfloat{listing}{tb}{lst}{}
\floatname{listing}{Listing}
\title{SinSEMI: A One-Shot Image Generation Model and \\ Data-Efficient Evaluation Framework for Semiconductor Inspection Equipment}
\author{
    ChunLiang Wu\textsuperscript{\rm 1},
    Xiaochun Li\textsuperscript{\rm 1} \\
}
\affiliations{
    \textsuperscript{\rm 1}Brightest Technology Inc \\

    josh.wu@brightest-tech.com, xiaochun.li@brightest-tech.com
}

\usepackage{bibentry}

\begin{document}

\maketitle

\begin{abstract}
In the early stages of semiconductor equipment development, obtaining large quantities of raw optical images poses a significant challenge. This data scarcity hinder the advancement of AI-powered solutions in semiconductor manufacturing. To address this challenge, we introduce SinSEMI, a novel one-shot learning approach that generates diverse and highly realistic images from single optical image. SinSEMI employs a multi-scale flow-based model enhanced with LPIPS (Learned Perceptual Image Patch Similarity) energy guidance during sampling, ensuring both perceptual realism and output variety. We also introduce a comprehensive evaluation framework tailored for this application, which enables a thorough assessment using just two reference images. Through the evaluation against multiple one-shot generation techniques, we demonstrate SinSEMI's superior performance in visual quality, quantitative measures, and downstream tasks. Our experimental results demonstrate that SinSEMI-generated images achieve both high fidelity and meaningful diversity, making them suitable as training data for semiconductor AI applications. Code is available at \href{https://github.com/JoshWuuu/SinSEMI-main}{https://github.com/JoshWuuu/SinSEMI-main}.
\end{abstract}


\section{Introduction}
In the early stages of semiconductor equipment development, collecting real optical images presents a significant challenge. This scarcity can be attributed to data proprietary concerns within the industry, as well as the inherent lack of data that is typical when a new collection process has just begun. While the industry typically relies on optical simulation methods to generate synthetic data for semiconductor devices in the early stage, these simulations are computationally intensive and time-consuming \cite{Sullivan2000FDTD,Taflove2005FDTD,dey2024addressing}. These constraints create a significant bottleneck for developing AI solutions in the semiconductor industry, as the scarcity of data significantly impedes the development and validation of robust machine learning models for semiconductor equipment applications \cite{dey2024addressing}.

Recent AI-driven image generation \cite{Eigenschink2023SyntheticData,Kazerouni2023Diffusion,Gao2024Prediff,cachay2023dyffusion} require extensive training datasets of thousands of images, making them impractical for semiconductor manufacturing where optical image data is severely limited in the preliminary stage. Several methods have been proposed for image synthesis under limited data, including data-efficient learning \cite{karras2020ADA, Liu2020fastGAN, mo2020freezeD}, few-shot learning\cite{gu2021lofgan, hong2022delta, Zhu2022FewShotDiffusion}, and one-shot learning approaches \cite{Shaham2019SinGAN, Kulikov2023SinDDM, Elnekave2022GPDM}. However, data-efficient learning struggles with the domain gap between natural and optical images, while few-shot learning requires more samples and categories than our available images  \cite{Yang2023LimitedData}. One-shot image generation, which learns to generate new images from a single example, emerges as the most suitable solution for our scenario since we have a very limited set of around 10 simulated images per structure type. This approach has shown great success in domains where data scarcity is an issue \cite{Chen2020oneshotMRI}. In addition, even when data-efficient AI solutions are proposed, validating their practical utility in he absence of sufficient real data remains a critical challenge. 

In this study, we introduce SinSEMI, a novel multi-scale flow-based model utilizing the LPIPS energy guidance for one-shot optical image generation in semiconductor applications. We propose a comprehensive and highly data-efficient evaluation framework for this task, which requires only two reference images to establish quantitative baselines and downstream test sets. This framework is comprehensive enough for semiconductor applications, combining visual assessment, quantitative metrics, and downstream task performance. Our results demonstrate that SinSEMI with energy guidance not only outperforms existing one-shot methods in generating high-quality optical images but also produces synthetic data that maintains its effectiveness in practical semiconductor AI applications.

The main contributions of this paper are as follows:
\begin{itemize}
    \item We propose SinSEMI, the first multi-scale flow-matching model for one-shot image generation specifically tailored for optical image generation in semiconductor applications.
    \item We incorporate a training-free mechanism, LPIPS energy guidance, to further enhance image quality during sampling.
    \item We develope a comprehensive and data-efficient evaluation framework that uses only two reference images to assess model performance across visual quality, quantitative metrics, and downstream tasks.
\end{itemize}

\section{Related Work}
\label{related_work}
\subsection{AI Approaches in the Semiconductor Industry}
In the semiconductor industry, significant research has been devoted to the classification, detection, and segmentation of defects using artificial intelligence (AI). Cheon et al. \cite{Cheon2019} developed an automatic defect classification (ADC) system utilizing electron microscope images. This system employs a single convolutional neural network (CNN) model designed to classify defects effectively. Experimental results on real datasets demonstrated the ADC system's strong classification performance. Dey et al. \cite{dey2022deep} proposed an innovative ensemble deep learning approach based on RetinaNet for the detection and localization of various defect categories. To enhance performance, predictions from multiple models are combined using ensemble methods. This approach notably improved the mean average precision (mAP), particularly for challenging defect classes. YOLOSeg \cite{li2025}, an end-to-end instance segmentation model, specifically addresses the segmentation of small particle defects. Experimental results indicate that YOLOSeg significantly outperforms other segmentation models in accuracy and reliability. Additionally, SEMI-DiffusionInst \cite{de2023semi} leverages a diffusion model for both defect detection and segmentation, showcasing its effectiveness in handling defect patterns.

Compared to defect identification and segmentation, research focused on image generation within the semiconductor industry remains relatively limited. YOLOSeg \cite{li2025} incorporates generative models primarily as a data augmentation strategy to diversify defect examples for training. Dey et al. \cite{dey2024addressing} introduced a diffusion model capable of synthesizing realistic semiconductor scanning electron microscope (SEM) images, specifically under conditions of limited available data. The synthetic images generated by this method closely resemble authentic SEM images and have successfully been integrated into training sets for defect detection tasks.

Despite the promise of image generation for addressing semiconductor data scarcity, current research is limited, especially for complex optical images.  Optical image generation presents a harder task due to their noisier and more complex nature compared to SEM images. In addition, the notable absence of one-shot generation studies, critical for extreme data limitations, underscores the urgent need for generative AI to synthesize diverse defect data in this field.



\subsection{One-Shot Image Generation}
One-shot image generation aims to learn the internal distribution and patterns from a single high-quality example to generate diverse new samples. It has three main categories, Generative Adversarial Networks (GANs), Diffusion Models and Non-Parametric Methods.
\subsubsection{Generative Adversarial Networks (GANs)}
GANs \cite{Goodfellow2014GAN} train two models in an adversarial setup for image generation: a generator that produces images from random noise and a discriminator that distinguishes real images from generated ones. SinGAN \cite{Shaham2019SinGAN} pioneered multi-stage GANs for one-shot image generation using a pyramid of generators and discriminators at different scales. Generation proceeds from the coarsest to finest scale, with each stage's output serving as input to the next stage. This hierarchical approach captures global layout at coarse scales while preserving fine details at finer scales. ConSinGAN \cite{Hinz2021ConSinGAN} improve SinGAN by introducing concurrent multi-stage GAN training. The stages were trained simultaneously with different learning rates, leading to enhanced image quality and reduced training time. Additionally, ConSinGAN introduced an improved image rescaling technique that enabled more training steps at lower resolutions, resulting in better global coherence in the generated images.
\subsubsection{Diffusion Models}
Diffusion models \cite{Sohl2015Diffusion,Ho2020DDPM} are probabilistic generative models that learn to reverse a gradual noise-adding process to generate images. SinDiffusion \cite{Wang2022SinDiffusion} pioneered the adaptation of diffusion models for single image generation. Unlike the multi-stage training approach of ConSinGAN, SinDiffusion employs a single diffusion model trained in a single stage. It prevents memorization by removing downsampling, upsampling, and attention layers, allowing its simplified architecture to focus on patch-wise statistics for more diverse outputs. SinFusion \cite{Nikankin2023SinFusion} adapts diffusion models for one-shot generation by training on large random crops (95\% of image size) rather than using a multi-scale pyramid, enabling preservation of global structure while introducing variations. SinDDM \cite{Kulikov2023SinDDM} adapts the diffusion model framework to single-image generation with multi-stage training. Like SinGAN and ConSinGAN, SinDDM employs a hierarchical approach where the denoising network is trained on different scales of the image, allowing it to generate samples in a coarse-to-fine manner.  
\subsubsection{Non-Parametric Methods}
Non-parametric methods for image generation operate by directly matching the distribution of image patches, rather than training parametric models like GANs or diffusion models. GPNN \cite{Granot2021GPNN} employs a hierarchical patch-based nearest-neighbor search to generate novel images. Operating in a coarse-to-fine manner inspired by SinGAN, it matches and rearranges patches across multiple scales to synthesize new images while preserving the patch distribution of the source image.
GPDM \cite{Elnekave2022GPDM} leverages the Sliced Wasserstein Distance (SWD) to efficiently match patch distributions between input and generated images. By projecting distributions onto one-dimensional subspaces and using random projections, GPDM achieves computationally efficient patch distribution matching without requiring adversarial training.

\section{Method}
\label{method}
\usetikzlibrary{shapes.geometric, arrows}
\tikzstyle{arrow} = [thick,->,>=stealth]
\tikzstyle{straight} = [thick,-,>=stealth]
\tikzstyle{dashedarrow} = [thick,dashed,->,>=stealth]
\tikzstyle{dottedline} = [thick, dotted, dash pattern=on 3pt off 2pt]
\tikzstyle{circlearrow} = [circle, draw, minimum size=0.4cm]
\begin{figure*}[ht]
\centering
\vskip 0.2in
\begin{tikzpicture}[node distance=0.7cm]

\draw[thick, blue!30, fill=blue!20, rounded corners=5pt] (7.7, 7.7) rectangle (15.7, 1.2);
\draw[thick, orange!30, fill=orange!20, rounded corners=5pt] (7.65, 7.7) rectangle (-1.5, 1.2);

\node at (3.5, 7.35) {\textbf{Multi-scale forward diffusion}};
\node at (12.0, 7.35) {\textbf{Multi-scale backward generation}};

\draw[thick, orange!10, fill=orange!10] (-0.3, 5.2) rectangle (7.6, 6.8); 

\draw[thick, orange!10, fill=orange!10] (-0.15, 2.85) rectangle (7.45, 4.15); 

\draw[thick, orange!10, fill=orange!10] (0, 1.3) rectangle (7.3, 2.3); 

\draw[thick, blue!10, fill=blue!10] (7.75, 5.2) rectangle (15.65 , 6.8); 

\draw[thick, blue!10, fill=blue!10] (7.9, 2.85) rectangle (15.5, 4.15); 

\draw[thick, blue!10, fill=blue!10] (8.05, 1.3) rectangle (15.35, 2.3);

\node at (-0.85, 6) {\fontsize{7}{7}\selectfont \textbf{Scale=N-1}};
\node at (-0.8, 3.5) {\fontsize{7}{7}\selectfont \textbf{Scale=1}};
\node at (-0.8, 1.8) {\fontsize{7}{7}\selectfont \textbf{Scale=0}};

\node at (0.5, 6.95) {\fontsize{7}{7}\selectfont \textbf{t = 0}};
\node at (2.6, 6.95) {\fontsize{7}{7}\selectfont \textbf{t}};
\node at (4.7, 6.95) {\fontsize{7}{7}\selectfont \textbf{t}};
\node at (6.8, 6.95) {\fontsize{7}{7}\selectfont \textbf{t = 1}};

\node at (8.55, 6.95) {\fontsize{7}{7}\selectfont \textbf{t = 1}};
\node at (10.65, 6.95) {\fontsize{7}{7}\selectfont \textbf{t = 1-h}};
\node at (12.75, 6.95) {\fontsize{7}{7}\selectfont \textbf{t = 1-2h}};
\node at (14.85, 6.95) {\fontsize{7}{7}\selectfont \textbf{t = 0}};

\node (A1) at (0.5,6) [draw, line width=1mm,, orange, inner sep=0.5pt] {\includegraphics[width=1.4cm]{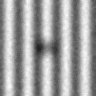}};
\node at (0.5,6.5) [text=orange, align=center] {\scriptsize Train \scriptsize Image};
\node (A2) [right of=A1, xshift=1.4cm] [draw, line width=0.5mm,, black, inner sep=0.3pt]{\includegraphics[width=1.4cm]{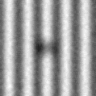}};
\node (A3) [right of=A2, xshift=1.4cm] [draw, line width=0.5mm,, black, inner sep=0.3pt]{\includegraphics[width=1.4cm]{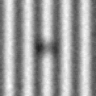}};
\node (A4) [right of=A3, xshift=1.4cm] [draw, line width=0.5mm,, black, inner sep=0.3pt]{\includegraphics[width=1.4cm]{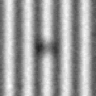}};

\draw [arrow] (1.3, 6.0) -- (1.8, 6.0);
\draw [arrow] (3.4, 6.0) -- (3.9, 6.0);
\draw [dashedarrow] (5.5, 6.0) -- (6.0, 6.0);

\node (B1) [below of=A1, yshift=-1.8cm] [draw, line width=0.5mm,, black, inner sep=0.3pt]{\includegraphics[width=1.1cm]{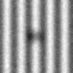}};
\node (B2) [right of=B1, xshift=1.4cm] [draw, line width=0.5mm,, black, inner sep=0.3pt]{\includegraphics[width=1.1cm]{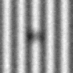}};
\node (B3) [right of=B2, xshift=1.4cm] [draw, line width=0.5mm,, black, inner sep=0.3pt]{\includegraphics[width=1.1cm]{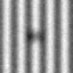}};
\node (B4) [right of=B3, xshift=1.4cm] [draw, line width=0.5mm,, black, inner sep=0.3pt]{\includegraphics[width=1.1cm]{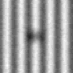}};

\draw [arrow] (1.2, 3.5) -- (1.9, 3.5);
\draw [arrow] (3.3, 3.5) -- (4.0, 3.5);
\draw [dashedarrow] (5.4, 3.5) -- (6.1, 3.5);

\node (C1) [below of=B1, yshift=-1.0cm] [draw, line width=0.5mm,, black, inner sep=0.3pt]{\includegraphics[width=0.8cm]{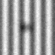}};
\node (C2) [right of=C1, xshift=1.4cm] [draw, line width=0.5mm,, black, inner sep=0.3pt]{\includegraphics[width=0.8cm]{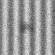}};
\node (C3) [right of=C2, xshift=1.4cm] [draw, line width=0.5mm,, black, inner sep=0.3pt]{\includegraphics[width=0.8cm]{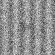}};
\node (C4) [right of=C3, xshift=1.4cm] [draw, line width=0.5mm,, black, inner sep=0.3pt]{\includegraphics[width=0.8cm]{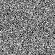}};

\draw [arrow] (1.1, 1.8) -- (2.0, 1.8);
\draw [arrow] (3.2, 1.8) -- (4.1, 1.8);
\draw [dashedarrow] (5.3, 1.8) -- (6.2, 1.8);

\draw[dottedline] (0.5, 5.3) -- ++(0, -0.5);
\draw[dottedline] (6.8, 5.3) -- ++(0, -0.5);

\draw [straight] (0.5, 4.8) -- ++(0, -0.75);
\draw [straight] (0.5, 2.95) -- ++(0, -0.75);

\draw [arrow] (0.8, 2.2) -- ++(0,0.375) -- ++(6,0) -- ++(0,0.375);

\node (circ1) [circlearrow, fill=white] at (0.5, 2.575) {};
\draw [<-, line width=0.3mm] (0.5, 2.425) -- ++(0,0.3);

\node (circ1) [circlearrow, fill=white] at (3.65, 2.575) {};
\draw [->, line width=0.3mm] (3.65, 2.425) -- ++(0,0.3);

\draw [arrow] (0.8, 4.05) -- ++(0,0.375) -- ++(6,0) -- ++(0,0.375);

\node (circ1) [circlearrow, fill=white] at (0.5, 4.425) {};
\draw [<-, line width=0.3mm] (0.5, 4.275) -- ++(0,0.3);

\node (circ1) [circlearrow, fill=white] at (3.65, 4.425) {};
\draw [->, line width=0.3mm] (3.65, 4.275) -- ++(0,0.3);

\node (D1) [right of=A4, xshift=1.05cm] [draw, line width=0.5mm,, black, inner sep=0.3pt]{\includegraphics[width=1.4cm]{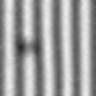}};
\node (D2) [right of=D1, xshift=1.4cm] [draw, line width=0.5mm,, black, inner sep=0.3pt]{\includegraphics[width=1.4cm]{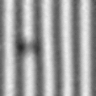}};
\node (D3) [right of=D2, xshift=1.4cm] [draw, line width=0.5mm,, black, inner sep=0.3pt]{\includegraphics[width=1.4cm]{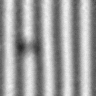}};
\node (D4) [right of=D3, xshift=1.4cm] [draw, line width=1mm,, blue, inner sep=0.5pt] {\includegraphics[width=1.4cm]{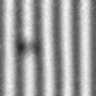}};
\node at (14.85,6.5) [text=blue, align=center] {\scriptsize Generation};

\draw [arrow] (9.35, 6.0) -- (9.85, 6.0);
\draw [arrow]  (11.45, 6.0) -- (11.95, 6.0);
\draw [dashedarrow] (13.55, 6.0) -- (14.05, 6.0);

\node (E1) [below of=D1, yshift=-1.8cm] [draw, line width=0.5mm,, black, inner sep=0.3pt]{\includegraphics[width=1.1cm]{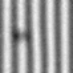}};
\node (E2) [right of=E1, xshift=1.4cm] [draw, line width=0.5mm,, black, inner sep=0.3pt]{\includegraphics[width=1.1cm]{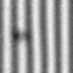}};
\node (E3) [right of=E2, xshift=1.4cm] [draw, line width=0.5mm,, black, inner sep=0.3pt]{\includegraphics[width=1.1cm]{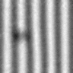}};
\node (E4) [right of=E3, xshift=1.4cm] [draw, line width=0.5mm,, black, inner sep=0.3pt]{\includegraphics[width=1.1cm]{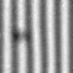}};

\draw [arrow] (9.25, 3.5) -- (9.95, 3.5);
\draw [arrow] (11.35, 3.5) -- (12.05, 3.5);
\draw [dashedarrow] (13.45, 3.5) -- (14.15, 3.5);

\node (F1) [below of=E1, yshift=-1.0cm] [draw, line width=0.5mm,, black, inner sep=0.3pt]{\includegraphics[width=0.8cm]{AnonymousSubmission/LaTeX/figs/4.png}};
\node (F2) [right of=F1, xshift=1.4cm] [draw, line width=0.5mm,, black, inner sep=0.3pt]{\includegraphics[width=0.8cm]{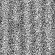}};
\node (F3) [right of=F2, xshift=1.4cm] [draw, line width=0.5mm,, black, inner sep=0.3pt]{\includegraphics[width=0.8cm]{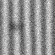}};
\node (F4) [right of=F3, xshift=1.4cm] [draw, line width=0.5mm,, black, inner sep=0.3pt]{\includegraphics[width=0.8cm]{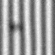}};

\draw [arrow] (9.15, 1.8) -- (10.05, 1.8);
\draw [arrow] (11.25, 1.8) -- (12.15, 1.8);
\draw [dashedarrow] (13.35, 1.8) -- (14.25, 1.8);

\draw[dottedline] (8.55, 5.3) -- ++(0, -0.5);

\draw [arrow] (14.85, 4.05) -- ++(0,0.375) -- ++(-6.3,0) -- ++(0,0.375);

\node (circ1) [circlearrow, fill=white] at (11.7,  4.425) {};
\draw [->, line width=0.3mm] (11.7, 4.275) -- ++(0,0.3);

\draw [arrow] (14.85, 2.2) -- ++(0,0.375) -- ++(-6.3,0) -- ++(0,0.375);

\node (circ1) [circlearrow, fill=white] at (11.7, 2.575) {};
\draw [->, line width=0.3mm] (11.7, 2.425) -- ++(0,0.3);


\end{tikzpicture}
\caption{Overview of SinSEMI's forward and backward processes: The forward process (left) progressively diffuses the optical image from the finest scale (Scale=N-1) to the coarsest scale (Scale=0) while building a pyramid of downsampled images, whereas the backward process (right) generates images in a coarse-to-fine manner through iterative denoising starting from Gaussian noise.}
\label{fig:forward_diffusion}
\vskip -0.2in
\end{figure*}

This section introduces SinSEMI, a novel multi-scale flow generative model designed for one-shot optical image generation in semiconductor applications. SinSEMI utilizes the Image-to-Image Schrödinger Bridge Conditional Flow Matching (SB-CFM) framework \cite{tong2023improving}. To further enhance sampling quality, it incorporates LPIPS energy guidance during the sampling process, ultimately aiming to produce high-fidelity optical images.

\subsection{SinSEMI Training}
The forward process of SinSEMI follows a multi-scale flow-based framework, as illustrated in the left panel of \cref{fig:forward_diffusion}. Given an input optical image $x^{N-1}_0$, we first construct a pyramid of downsampled images $x^{s}_0$ using bilinear interpolation, where s denotes the scale level and ranges from $N-1$ to $0$. At each scale $s$, we define the forward diffusion process that gradually transforms $x^s_0$ into a target state $x^s_1$. For intermediate scales $s < N-1$, the target state $x^s_1$ is derived from the upsampled version of the image at the next coarser scale $x^{s-1}_0$. At the final scale $s = 0$, the target state $x^0_1$ is set to be Gaussian noise $\mathcal{N}(0, I)$, which introduces additional variation during the sampling process.

The conditional distribution of \(x_t^s\) given \(x_0^s\) and \(x_1^s\) is defined as:
\begin{equation}
\label{eq:forward}
\begin{aligned}
p_t^s(x \mid x_0^s, x_1^s) = \mathcal{N}(x \mid (1-t) x_0^s + t x_1^s, t(1-t)\sigma^2)
\end{aligned}
\end{equation}

The corresponding velocity (vector field) at time \(t\) is obtained by taking the partial derivative of the intermediate state \(x_t^s\) with respect to \(t\):

\begin{equation}
\begin{aligned}
\label{eq:2}
u_t^s(x \mid x_0^s, x_1^s) &= x_1^s - x_0^s + \sigma \cdot c1, \\
\text{where} \quad c1 &= \frac{1 - 2t}{2\sqrt{t(1-t)} + \epsilon} \\
\end{aligned}
\end{equation}

The model, parameterized by \(\theta\), predicts the velocity field \(u_\theta(x_t^s, t, s)\) conditioned on timestep \(t\) and scale \(s\). The Conditional Flow Matching (CFM) loss is then defined as the mean squared error between the predicted and true velocity fields:
\begin{equation}
\label{eq:cfm_loss}
\mathcal{L}_{\text{CFM}} = \mathbb{E}_{x_0^s, x_1^s, s, t}\left[\left\|u_\theta(x_t^s, t, s) - u_t^s(x \mid x_0^s, x_1^s)\right\|^2\right]
\end{equation}



\subsection{SinSEMI Sampling}
As illustrated in the right panel of \cref{fig:forward_diffusion}, the backward generation process of SinSEMI follows a coarse-to-fine strategy across multiple scales. The process begins at the coarsest scale ($s=0$) with Gaussian noise sampled from $\mathcal{N}(0, I)$. At each scale \(s\), we apply the Euler–Maruyama discretization of Stochastic Differential Equation (SDE) to progressively convert the noisy signal into a denoised image. The resulting image $\hat{x}_0^s$ is then upsampled to the next finer scale using bilinear interpolation, serving as the initial state $\hat{x}_T^{s+1}$ for the subsequent denoising process. This procedure continues until reaching the finest scale $\hat{x}_0^{N-1}$.

Specifically, given the trained velocity field \(u_\theta\), the intermediate state \(\hat{x}_t^s\) evolves according to:
\begin{equation}
\frac{d \hat{x}_t^s}{d t} = u_\theta(\hat{x}_t^s, t, s)
\end{equation}

Moreover, we leverage the training-free guidance framework of Yu et al. \cite{yu2023freedom} to more effectively steer the generation process via \(\nabla_{\hat{x}_t^s}\log p(\hat{x}_t^s \mid c)\). To promote perceptual fidelity, we adopt LPIPS (Learned Perceptual Image Patch Similarity) \cite{Zhang2018Perceptual} as our energy function \(\mathcal{E}(c, \hat{x}_t^s)\). Concretely, we diffuse the conditional image \(c\) from the training images \(x_0^s\) and \(x_1^s\) according to \eqref{eq:forward}, then use this perturbed reference to compute the LPIPS distance against the generated output \(\hat{x}_t^s\). Finally, we inject the negative gradient of this energy into the Euler–Maruyama update alongside the model’s score estimate.

\begin{equation}
\begin{aligned}
&\nabla_{\hat{x}_t^s}  \log p(\hat{x}_t^s\mid c) \\
&\approx -\,\nabla_{\hat{x}_t^s}\mathcal{E}(c,\hat{x}_t^s) = \space -\,\nabla_{\hat{x}_t^s}\mathrm{LPIPS}\bigl(p^s_t(c \mid x_0^s, x^s_1),\,\hat{x}_t^s\bigr)
\end{aligned}
\end{equation}

The complete implementation details of the generation process are provided in~\ref{alg:sample}.

\begin{algorithm}[tb]
   \caption{SinSEMI Generation}
   \label{alg:sample}
\begin{algorithmic}
  \STATE Set step size $h = \frac{1}{n}$
  \FOR{$s = 0, \dots, N-1$}
    \IF{$s = 0$}
      \STATE $x_T^s \sim \mathcal{N}(0, I)$
    \ENDIF
    \STATE Set $t = 1$
    \FOR{$i = 0, \dots, n$}
        \STATE Gaussian noise $\epsilon_c \sim \mathcal{N}(0, I)$
        \STATE $x_t^s = (1-t) x_0^s + t x_1^s + \sqrt{t(1-t)} \sigma \epsilon_c$
        \STATE $g = -\,\nabla_{\hat{x}_t^s}\mathrm{LPIPS}\bigl(x_t^s,\,\hat{x}_t^s\bigr)$
        \STATE Gaussian noise $\epsilon \sim \mathcal{N}(0, I)$
        \STATE $x_{t-h} = x_t - h(u(x_t^s, t, s; \theta) - g) + \sqrt{h}\sqrt{t(1-t)} \sigma\epsilon$
        \STATE $t = t - h$
    \ENDFOR
  \ENDFOR
\end{algorithmic}
\end{algorithm}

\subsection{Model Structures}
SinSEMI adopts the model structure design from SinDDM \cite{Kulikov2023SinDDM}. The model is conditioned on both the scale $s$ and the timestep $t$. The architecture consists of four convolutional blocks, with a total receptive field of 35 by 35. This small receptive field prevents the model from memorizing the structure of the single input image. The same architectural design allows SinSEMI to directly compare the training and sampling methods between general denoising diffusion and the SB-CFM in a multi-scale one-shot generation framework. 

\subsection{Comparison to SinDDM}
Inspired by the multi‐scale architecture of SinDDM \cite{Kulikov2023SinDDM}, SinSEMI extends SB-CFM framework of Tong et al. \cite{tong2023improving} to semiconductor image synthesis. Rather than mapping clean images to Gaussian noise, SB-CFM treats each scale‐wise transformation as an entropy-regularized optimal transport problem between clean and degraded image pairs. This entropic regularization yields smoother transport maps that mitigate interpolation artifacts and preserves high-frequency details critical for accurate optical image generation. A more detailed comparison of the two methods can be found in the following section. 

\section{Experiments}
\label{experiments}

\subsection{Experimental Setup}
\begin{figure}[t]
\begin{center}
    \centering
    \includegraphics[width=0.7\linewidth]{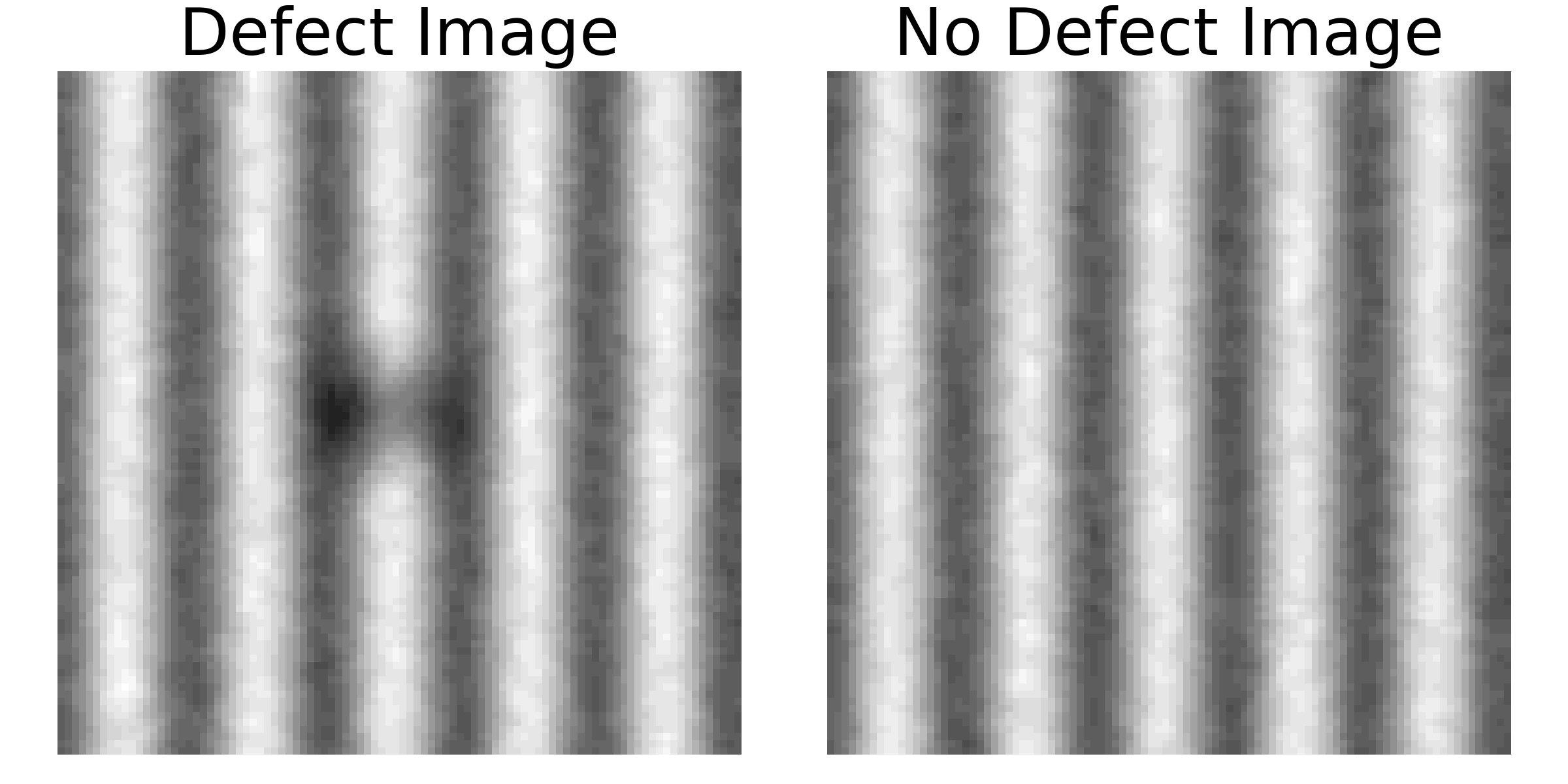}
    \caption{Optical simulation images of line pairs. The left image shows a line pair with a bridge defect, while the right image shows a line pair without defects}
    \label{fig:defect comparison}
\end{center}
\end{figure}

\paragraph{Datasets:} Each model is trained using a single 96 × 96 pixel optical simulation image of a line pair structure, a common feature in semiconductor chips, specifically focusing on line pairs with bridge defects (as illustrated in the left plot of \cref{fig:defect comparison}). The choice of line pairs is crucial for assessing the model's ability to preserve precise patterns and spatial relationships with a noisy background. Success on this common and challenging pattern provides a strong proof-of-concept for the SinSEMI framework's potential in semiconductor manufacturing.

\paragraph{Hyperparameters:} SinSEMI was trained with a scale factor of 1.4 for the multi-scale architecture. The $\sigma$ value for training and sampling was 0.35 for line pairs. Inference used 300 timesteps per scale. Training is performed with a batch size of 32 for 120,000 iterations using a learning rate of $1e-4$.

Beyond structural and noise generation, our evaluation primarily assesses the models' ability to faithfully reproduce and reasonably vary defect characteristics within the line pair structure. In the highly controlled application of semiconductor inspection, real-world variability is not arbitrary but is constrained to realistic changes in defect count, defect location, and subtle structural deformations. The underlying structural patterns and noise characteristics are expected to remain consistent. An ideal model must therefore generate defects that can maintain physical integrity and create useful variations—such as additional defects, defects in novel locations, or lines with realistic roughness—to better simulate real-world variations.

To assess these qualities, we conducted three experiments: (1) Visual Inspection of generated samples, (2) Quantitative Evaluation using SIFID and LPIPS, and (3) Defect Segmentation to evaluate image quality. All evaluation tests were conducted using 1000 generated images from each model.

\subsection{Visual Inspection}
\begin{figure*}[t]
\vskip 0.2in
\begin{center}
    \centering
    \includegraphics[width=1.0\linewidth]{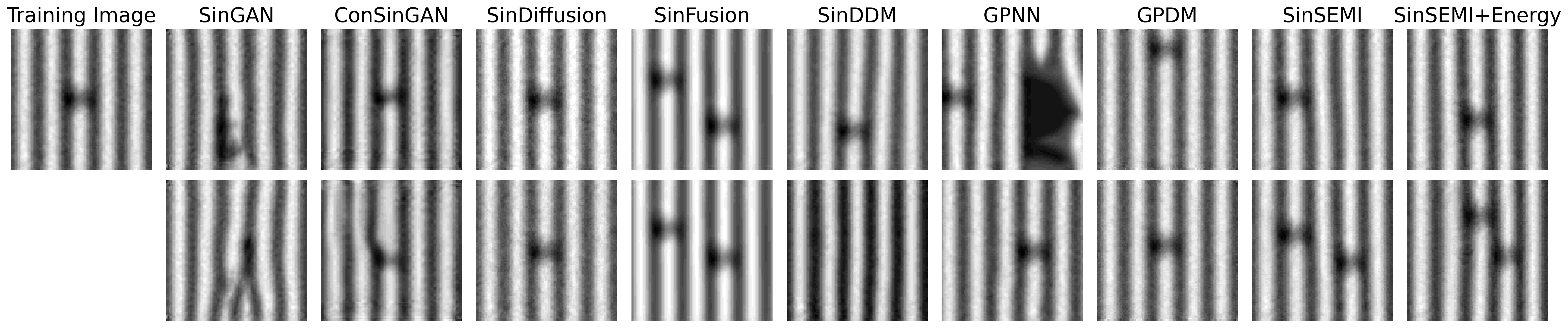}
    \caption{Visual comparison of generated samples across different generative models for line pairs. The training image for each structure is shown on the left, followed by samples generated by different models.}
    \label{fig:all_images}
\end{center}
\vskip -0.2in
\end{figure*}

\begin{figure*}[t]
\vskip 0.2in

    \centering
    \includegraphics[width=1.0\linewidth]{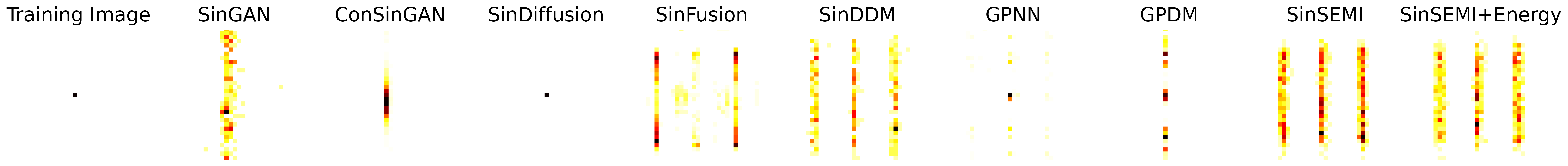}
    \caption{Spatial distribution of defects generated by different generative models for line pairs. Heat maps show accumulated defect locations from 1000 generated samples per model, where darker colors indicate higher frequency of defects at each location.}
    \label{fig:all_defects}
\vskip -0.2in
\end{figure*}
\begin{table}[t]
\caption{Line pairs' average defect counts for different models}
\label{tab:defect count results}
\begin{center}
\begin{small}
\begin{sc}
\begin{tabular}{ lc}
\toprule
Algorithms & Aerage defect count \\ \midrule
SinGAN       & 0.3    \\
ConSinGAN    & 1   \\ 
SinDiffusion  & 1  \\
SinFusion     & 1.4   \\ 
SinDDM    & 0.4    \\ 
GPNN          & 0.8   \\ 
GPDM       & 0.8   \\ 
\textbf{SinSEMI} & \textbf{0.8} \\
\textbf{SinSEMI+energy} & \textbf{0.8} \\
\bottomrule
\end{tabular}
\end{sc}
\end{small}
\end{center}
\end{table}
We first visually inspected the generated line pair samples across various model architectures (\cref{fig:all_images}). To complement this, \cref{fig:all_defects} shows the spatial distribution of generated defects for each method. \cref{tab:defect count results} presents the average defect count per model. Counts under 0.5 are too low for effective downstream defect segmentation (indicating a lack of defect learning), while a count of 1 often signals overfitting. A comprehensive visual inspection requires combining insights from these two figures and the table.

Both SinGAN and ConSinGAN produced unrealistic structural line deformations. SinGAN's defects were mostly confined to the middle line, and its low average defect count was also problematic for downstream tasks, as it offered insufficient defect patterns for effective learning. ConSinGAN's defects were concentrated in the middle and had an average defect count of exactly 1, indicating strong overfitting. SinDiffusion exhibited clear overfitting, evidenced by a defect count of 1, consistent central defect spatial distribution, and highly similar image quality across samples. SinFusion produced overly smooth images, demonstrating poor learning of the noisy background. SinDDM generated too few defects, with approximately half of its images resembling the defect-free examples. Both GPNN and GPDM struggled with defect location overfitting. Despite this, both methods generated good image quality, particularly regarding the noisy background.

SinSEMI and its energy-guided variant achieved superior image quality. They effectively addressed the limitations of prior models, such as the deformations seen in SinGAN and ConSinGAN, and the oversmoothness encountered with SinFusion. Furthermore, unlike models hindered by overfitting (SinDiffusion, ConSinGAN, GPNN, GPDM) or insufficient defect generation (SinGAN, SinDDM), SinSEMI consistently generated highly varied defect locations and maintained optimal average defect counts. This combination of high-fidelity image generation and diverse and realistic defect characteristics positions SinSEMI as a significant advancement in the field.

\subsection{Quantitative Evaluation - SIFID and LPIPS
} 

\begin{table*}[t]
\caption{SIFID (×$10^3$) and LPIPS (×$10^3$) scores for different model architectures evaluated on line pair structures.}
\label{tab:quantitative results}
\begin{center}
\begin{small}
\begin{sc}
\begin{tabular}{ l cc }
\toprule
Algorithms & SIFID($\pm$ std) & LPIPS($\pm$ std) \\ 
\midrule
Diff between two optical images 
               & 2.4   & 0.8   \\ 
SinGAN       & 1.9($\pm$ 0.8)   & 1.6($\pm$ 0.7)   \\
ConSinGAN    & 1.3($\pm$ 0.6)  &  0.8($\pm$ 0.2)  \\ 
SinDiffusion & 2.6($\pm$ 0.4)  & 0.5($\pm$ 0.2)  \\
SinFusion     & 6.9($\pm$ 4.4)   &  3.2($\pm$ 0.7)   \\ 
SinDDM       & 2.2($\pm$ 1.1)   & 1.1($\pm$ 0.3)   \\ 
GPNN         & 6.0($\pm$ 8.9)  & 2.4($\pm$ 3.1)  \\ 
GPDM         & 2.1($\pm$ 1.2)   & 0.8($\pm$ 0.4) \\ 
\textbf{SinSEMI}       & \textbf{2.6($\pm$ 1.0)}   & \textbf{1.0($\pm$ 0.3)} \\
\textbf{SinSEMI+Energy}       & \textbf{2.4($\pm$ 0.9)}   & \textbf{0.8($\pm$ 0.3)} \\
\bottomrule
\end{tabular}
\end{sc}
\end{small}
\end{center}
\end{table*}

We evaluate image quality using two primary metrics: SIFID (Single Image Fréchet Inception Distance) \cite{Shaham2019SinGAN} to measure fidelity and LPIPS (Learned Perceptual Image Patch Similarity) \cite{Zhang2018Perceptual} to measure diversity. To ground our evaluation, we first establish a performance baseline by calculating these metrics between a pair of optical simulation images—one with a defect and one without—under identical conditions.

This comparison between a defect-free and a with-defect image establishes a quantitative baseline. The two reference images are nearly identical in their core structure, patterns, and noise characteristics, with the primary difference being the presence of the defect in the center (\cref{fig:defect comparison}). Our generative model is intended to replicate these core characteristics while introducing realistic defect variations. Consequently, the SIFID and LPIPS scores between the two optical simulation images represent an ideal target for this "meaningful difference". When a generated image's scores align with this baseline, it indicates the model has successfully produced a realistic variation rather than simply overfitting or introducing unrealistic artifacts.

Generated images were then computed against the optical simulation image containing a defect. Ideally, SIFID and LPIPS scores should closely approach the established baseline (SIFID = 2.4 and LPIPS = 0.8). Based on empirical observations across all experimental images, ideal SIFID scores fall within the 2.0 to 2.8 range, while LPIPS scores should be between 0.5 and 1.1. In addition, to measure the consistency of each model, the standard deviation of both metrics is also considered to assess performance reliability.

The quantitative results in \cref{tab:quantitative results} show that SinGAN's LPIPS of 1.6, exceeding the 1.1 threshold, indicated unrealistic variations. ConSinGAN achieved the ideal LPIPS score of 0.8, but its SIFID of 1.3 falled outside the desired range, suggesting it produced images that lack realistic details. SinDiffusion and SinDDM were competitive, though SinDiffusion has shown overfitting in visual inspection. Neither perfectly aligned with the baseline metrics. SinFusion performed poorly on both metrics. GPDM emerged as a very strong competitor, achieving a perfect LPIPS of 0.8 and a close SIFID of 2.1. In contrast, GPNN failed to generate plausible results. Furthermore, the low standard deviation in ConSinGAN and SinDiffusion is attributed to their overfitting behavior, leading to repetitive outputs.

Our proposed method, SinSEMI, demonstrated a strong balance, with a SIFID of 2.6 and an LPIPS of 1.0, both fell within the optimal ranges. It also achieves low standard deviations for both metrics (1.0 and 0.3), demonstrating its ability to produce consistently high-quality and reliable results. Crucially, the addition of training-free LPIPS energy guidance elevated the performance. SinSEMI+Energy emerged as the best model, perfectly matching the baseline SIFID of 2.4 and LPIPS of 0.8. This demonstrates our method's superior ability to generate images with both high fidelity and realistic diversity to the target optical characteristics.

\subsection{Defect Segmentation}
\label{segmentation}
To further assess the practical utility of the generated images for real-world applications, we use them in a downstream defect segmentation task, which is critical for quality control in the semiconductor industry. We trained a U-Net \cite{Ronneberger2015UNet} on 1000 images from each generation method and evaluated it on a test set of 50 optical simulation images, measuring performance with the Intersection over Union (IoU) metric. 

Our test set consists of 50 images, generated by applying noise and flipping augmentations to the two optical simulations shown in \cref{fig:defect comparison}. This augmentation strategy reflects realistic physical constraints for line pair structures and results in most test defects being centrally located.

While a 50-image set may seem insufficient for a comprehensive generalizability study, preliminary quantitative analysis showed statistically consistent results across test sets of 50, 100, and 500 images. This finding allowed us to proceed with the 50-image set to establish a "foundational standard": verifying that the synthetic data could effectively train a model to segment defects in their most common location. This foundational test is highly indicative of broader performance, as a model achieving high accuracy on the central case is expected to be well-suited for handling spatial variations (e.g., corner defects) within the highly similar structural patterns.

To establish a performance baseline for the downstream segmentation task, a U-Net model was trained on a dataset of 1000 images created by augmenting the two original optical simulation images. Due to the augmentation strategy, the defects in this baseline training set remained in the central location. This baseline serves as a critical benchmark for evaluating the one-shot generative models. Superior performance against this baseline indicates that a generator's synthetic images not only preserve structural consistency but also provide meaningful variations in defect location while preserving structural consistency, leading to more robust model training. Conversely, performance similar to or worse than the baseline suggests that the generated images contain unrealistic deformations that hinder effective training.

\begin{table}[t]
\caption{IoU for different models evaluated on line pair structures}
\label{tab:segmentation results}
\begin{center}
\begin{small}
\begin{sc}
\begin{tabular}{ lc}
\toprule
Algorithms & IoU ($\pm$ std) \\ \midrule
Baseline  & 0.74 ($\pm$ 0.41) \\
SinGAN        & 0.72 ($\pm$ 0.41)   \\
ConSinGAN     & 0.78  ($\pm$ 0.33) \\ 
SinDiffusion    & 0.79 ($\pm$ 0.28)  \\
SinFusion     & 0.76  ($\pm$ 0.3) \\ 
SinDDM    & 0.74   ($\pm$ 0.42) \\ 
GPNN          & 0.74  ($\pm$ 0.42) \\ 
GPDM        & 0.84  ($\pm$ 0.18) \\ 
\textbf{SinSEMI}& \textbf{0.82 ($\pm$ 0.26)} \\
\textbf{SinSEMI+energy} & \textbf{0.88 ($\pm$ 0.16)} \\
\bottomrule
\end{tabular}
\end{sc}
\end{small}
\end{center}
\end{table}

\begin{figure}[t]

    \centering
    \includegraphics[width=1\linewidth]{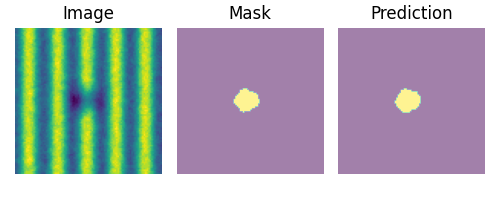}
    \caption{Defect segmentation results for SinSEMI+Energy. (Left) Testing optical image. (Middle) Ground truth defect segmentation. (Right) Predicted defect segmentation by the SinSEMI+Energy model.}
    \label{fig:seg line pair}
    
\end{figure}

As shown in \cref{tab:segmentation results}, SinSEMI with LPIPS energy guidance achieved the highest IoU of 0.88, followed by SinSEMI model with a strong IoU of 0.82. Crucially, these two models were the only methods to substantially outperform the baseline, excluding models like SinDiffusion, ConSinGAN and GPDM whose competitive scores are attributed to previously identified defect location overfitting. Other methods yielded IoU scores lower than the baseline, consistent with their known limitations; for instance, SinGAN's generation of problematic structural patterns hindered effective segmentation training.

Furthermore, the standard deviation offers insights into performance consistency. A high standard deviation ($\sim$ 0.4), as seen in the baseline and several other models, suggests inconsistent performance where a model sometimes fails to detect the defect entirely. In contrast, SinSEMI+Energy not only has the highest IoU but also the lowest standard deviation (0.16), confirming its superior reliability. 

\subsection{Discussion}
Our evaluation framework, including three distinct perspectives—visual quality, quantitative metrics (SIFID and LPIPS), and downstream task performance (segmentation IoU)—was crucial for a comprehensive data-efficient assessment of the models. This multifaceted approach allowed us to identify models that might perform good in one evaluation but falter in others. SinSEMI, particularly with LPIPS energy guidance, consistently excelled across all evaluations. It achieved the best possible scores for both fidelity (SIFID of 2.4) and diversity (LPIPS of 0.8), perfectly matching the real optical image baseline. Crucially, this performance translated directly into state-of-the-art results on the defect segmentation task, with an IoU of 0.88. Furthermore, it also achieved the lowest standard deviation among all metrics, demonstrating superior reliability. This comprehensive outcome affirms that our proposed method is not only capable of generating visually realistic data but is also functionally effective for robustly training downstream models in semiconductor inspection.

\section{Abalation}
\label{Abalation}

\begin{table}[t]
\caption{SIFID and LPIPS scores for different numbers of diffusion steps}
\label{tab:steps results}
\begin{center}
\begin{small}
\begin{sc}
\begin{tabular}{ l cc }
\toprule
Steps & SIFID & LPIPS \\ 
\midrule
10       & 2.5   & 0.7   \\
50    & 2.4  &  0.7  \\ 
100 & 2.4  & 0.8  \\
\bottomrule
\end{tabular}
\end{sc}
\end{small}
\end{center}
\end{table}
The results of our ablation study on the number of diffusion steps are presented in \cref{tab:steps results}. We observed that reducing the number of diffusion steps from 100 to 10 led to only a minor increase of 0.1 in the SIFID score (from 2.4 to 2.5), while the LPIPS score remained stable. It performs well within an acceptable range. Given that a lower number of steps corresponds to a significant reduction in computational cost and inference time, these results demonstrate an excellent trade-off between model performance and efficiency. 

\begin{table}[t]
\caption{SIFID (×$10^3$) and LPIPS (×$10^3$) scores for different LPIPS guidance strength}
\label{tab:guidance results}
\begin{center}
\begin{small}
\begin{sc}
\begin{tabular}{ c cc }
\toprule
LPIPS guidance strength & SIFID & LPIPS \\ 
\midrule
0      & 2.6   & 1.0   \\
0.1    & 2.6  &  0.9  \\ 
0.5 & 2.5  & 0.8  \\
1 & 2.4 & 0.8 \\
\bottomrule
\end{tabular}
\end{sc}
\end{small}
\end{center}
\end{table}
The influence of the LPIPS guidance strength on generation quality was presented in \cref{tab:guidance results}. We observed a direct correlation between the guidance strength and model performance. As the strength was increased from 0 to 1.0, the SIFID score progressively improved, decreasing from 2.6 to 2.4. Similarly, the LPIPS score saw a reduction from 1.0 to 0.8. This demonstrates that applying a stronger LPIPS guidance effectively enhances the perceptual quality of the generated images, with a strength of 1.0 providing the optimal results in our study.

\section{Future Direction}
\label{future direction}
For future work, rather than pursuing a single, generalized model that may compromise accuracy, we will adapt the SinSEMI framework to create specialized models for specific, critical semiconductor structures. This approach aligns with industry demand for highly optimized solutions and will involve tailoring the model to the unique characteristics of patterns such as contact holes and SRAM cells.

\section{Conclusion}
\label{conclusion}
In this work, we introduced SinSEMI, a novel framework to address critical data scarcity in semiconductor inspection. By leveraging a multi-scale, image-to-image generative model with LPIPS energy guidance, SinSEMI generates high-fidelity optical images with realistic defect characteristics. Using a data-efficient evaluation framework based on only two reference images, we confirmed its state-of-the-art performance across visual quality, quantitative metrics (SIFID/LPIPS), and a practical downstream segmentation task. SinSEMI provides a robust solution for synthetic data generation, enabling effective AI model training during early-stage semiconductor development when real data is unavailable.

\bibliography{aaai2026}

\end{document}